\documentclass[conference,letterpaper]{IEEEtran}
\IEEEoverridecommandlockouts

\usepackage{geometry}

\usepackage{cite}
\usepackage{svg}
\usepackage{amsmath,amssymb,amsfonts}
\usepackage{algorithm}
\usepackage{algorithmic}
\usepackage{graphicx}
\usepackage{textcomp}
\usepackage{xcolor}
\usepackage[numbers]{natbib}
\usepackage{booktabs}
\usepackage{multirow}
\usepackage{pdfpages}
\begin{document}

\title{Navigating High-Degree Heterogeneity: Federated Learning in Aerial and Space Networks}

\author{
\IEEEauthorblockN{Fan Dong\IEEEauthorrefmark{1}, Henry Leung\IEEEauthorrefmark{1}, Steve Drew\IEEEauthorrefmark{1}}
\IEEEauthorblockA{
\IEEEauthorrefmark{1}Department of Electrical and Software Engineering, University of Calgary, Calgary, AB, Canada \\
\{fan.dong, leungh, steve.drew\}@ucalgary.ca}
}

\maketitle

\begin{abstract}
Federated learning offers a compelling solution to the challenges of networking and data privacy within aerial and space networks by utilizing vast private edge data and computing capabilities accessible through drones, balloons, and satellites. While current research has focused on optimizing the learning process, computing efficiency, and minimizing communication overhead, the heterogeneity issue and class imbalance remain a significant barrier to rapid model convergence.
In this paper, we explore the influence of heterogeneity on class imbalance, which diminishes performance in Aerial and Space Networks (ASNs)-based federated learning. We illustrate the correlation between heterogeneity and class imbalance within grouped data and show how constraints such as battery life exacerbate the class imbalance challenge. Our findings indicate that ASNs-based FL faces heightened class imbalance issues even with similar levels of heterogeneity compared to other scenarios.
Finally, we analyze the impact of varying degrees of heterogeneity on FL training and evaluate the efficacy of current state-of-the-art algorithms under these conditions. Our results reveal that the heterogeneity challenge is more pronounced in ASNs-based federated learning and that prevailing algorithms often fail to effectively address high levels of heterogeneity.
\end{abstract}


\section{Introduction}
Aerial and Space Networks (ASNs) \citep{liu2018space} represent a novel type of network that integrates aerial and space assets, including drones, balloons, and satellites. These assets are interconnected, enabling the collection and relay of diverse sensing data across various-speed and universal data networks. Additionally, the computational capabilities of these assets in ASNs can facilitate edge computing, allowing for complex machine-learning tasks to be performed locally \citep{zhang2022aerial}.
However, the heterogeneous nature of these devices, limited bandwidth, and differing ownerships present significant challenges for data processing and the centralized training of predictive models in ASNs. The primary challenges include limited bandwidth, privacy concerns, and single-point failure.

Federated learning (FL) \citep{mcmahan2017communication} has emerged as a promising solution, where distributed clients train their models locally and only send the model parameters to a central server. However, data heterogeneity remains a significant challenge in this context. Data distributed across Internet of Things (IoT) devices and edge servers or nodes leads to each participant owning a unique local dataset. These datasets can vary significantly in size, feature space, and label distribution, resulting in discrepancies in local model performance, and consequently, in the aggregated global model's performance.

Moreover, data heterogeneity can slow down the convergence of FL. Variations in local data distributions can cause local models to diverge significantly, complicating the aggregation into a robust global model. Addressing data heterogeneity often requires more communication rounds between the central server and the nodes to achieve acceptable model performance, which incurs higher bandwidth usage, particularly costly in resource-constrained edge IoT environments with limited connectivity. Effectively managing data heterogeneity also necessitates more sophisticated methods to aggregate local models or to carefully train and adapt local models to diverse data distributions.

\begin{figure}[t]
    \centering
    \includegraphics[width=0.9\linewidth]{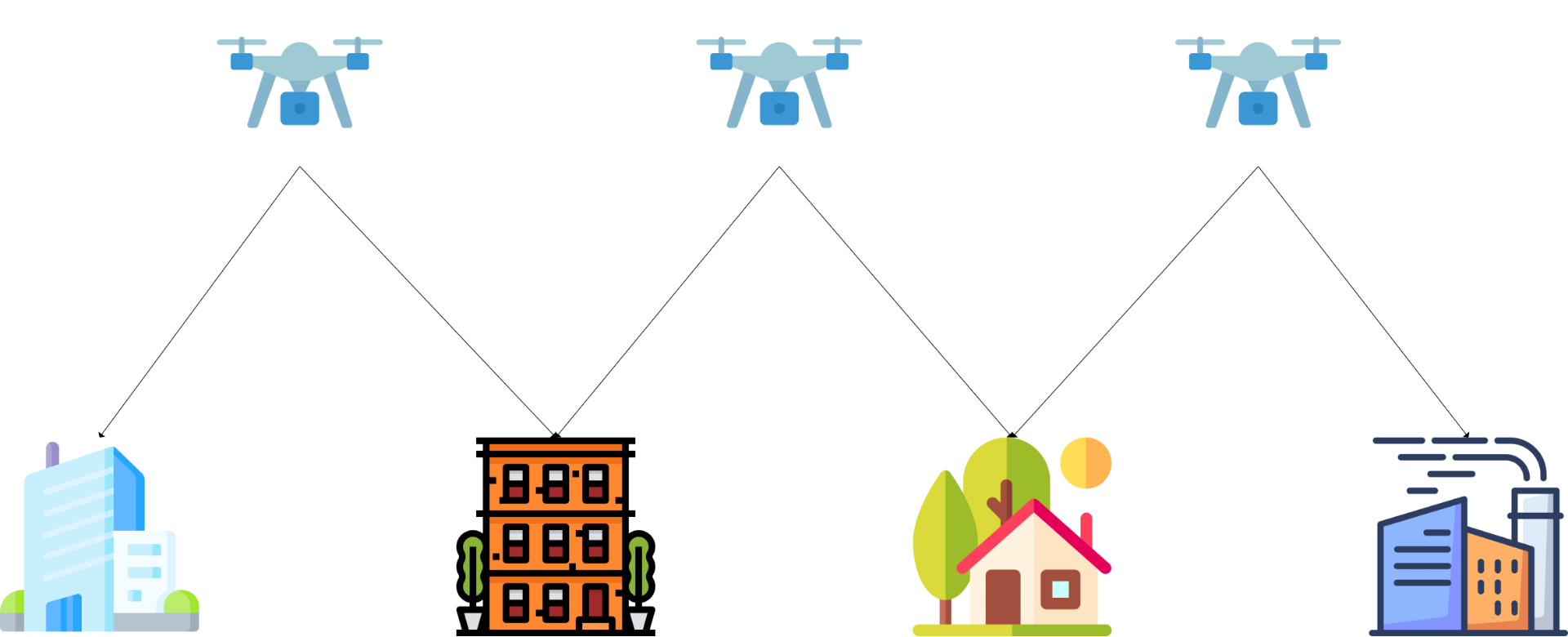}
    \caption{
    Different devices would possess various data concerning their active locations.}
    \label{fig:drones}
\end{figure}

Without exception, ASNs-based edge devices are designed to complete diverse tasks with various intensities, resulting in high data heterogeneity. For instance, in the low-altitude economy era, drones capture images of different types of buildings in the city. As shown in Fig. \ref{fig:drones}, drones will confront various surroundings like office buildings, residential buildings, houses, and factories. And corresponding flight strategies will also be developed to cope with these varying conditions. Due to the variety of the aviation environment, different drones will collect heterogeneous data accordingly.
These issues, combined with the existing uneven distribution of data, may further increase the severity of data heterogeneity due to
\paragraph{Statistical Heterogeneity}
Statistical heterogeneity occurs when the data distributions across different devices or nodes vary significantly. In FL settings, each ASNs node may collect data under different conditions, leading to non-independently and identically distributed (non-IID) data. This heterogeneity can lead to biased models that perform well on some nodes but poorly on others, as the global model might not generalize well across diverse datasets.

\paragraph{System Heterogeneity}
This refers to differences in hardware, network connectivity, and computational power among devices participating in federated learning. Some devices may be able to compute updates faster and more frequently than others. This discrepancy can lead to slower convergence of the global model, as updates from less capable devices might be received less frequently or could be outdated.

\paragraph{Communication Heterogeneity}
Variations in network speed and bandwidth across devices can affect the efficiency of data transmission in federated learning. Devices with slower network connections may take longer to upload their updates, leading to delays in model aggregation and potentially outdated model updates being incorporated into the global model.

\paragraph{Label Distribution Skew}
In some cases, the distribution of labels (outcomes of interest) can differ significantly across devices. For example, in a healthcare application, data collected from different hospitals might show different disease prevalence rates. This skew can lead to a model biased towards the data characteristics of more frequently represented labels or devices.

\begin{figure}[htbp]
    \centering
    \includegraphics[width=\linewidth]{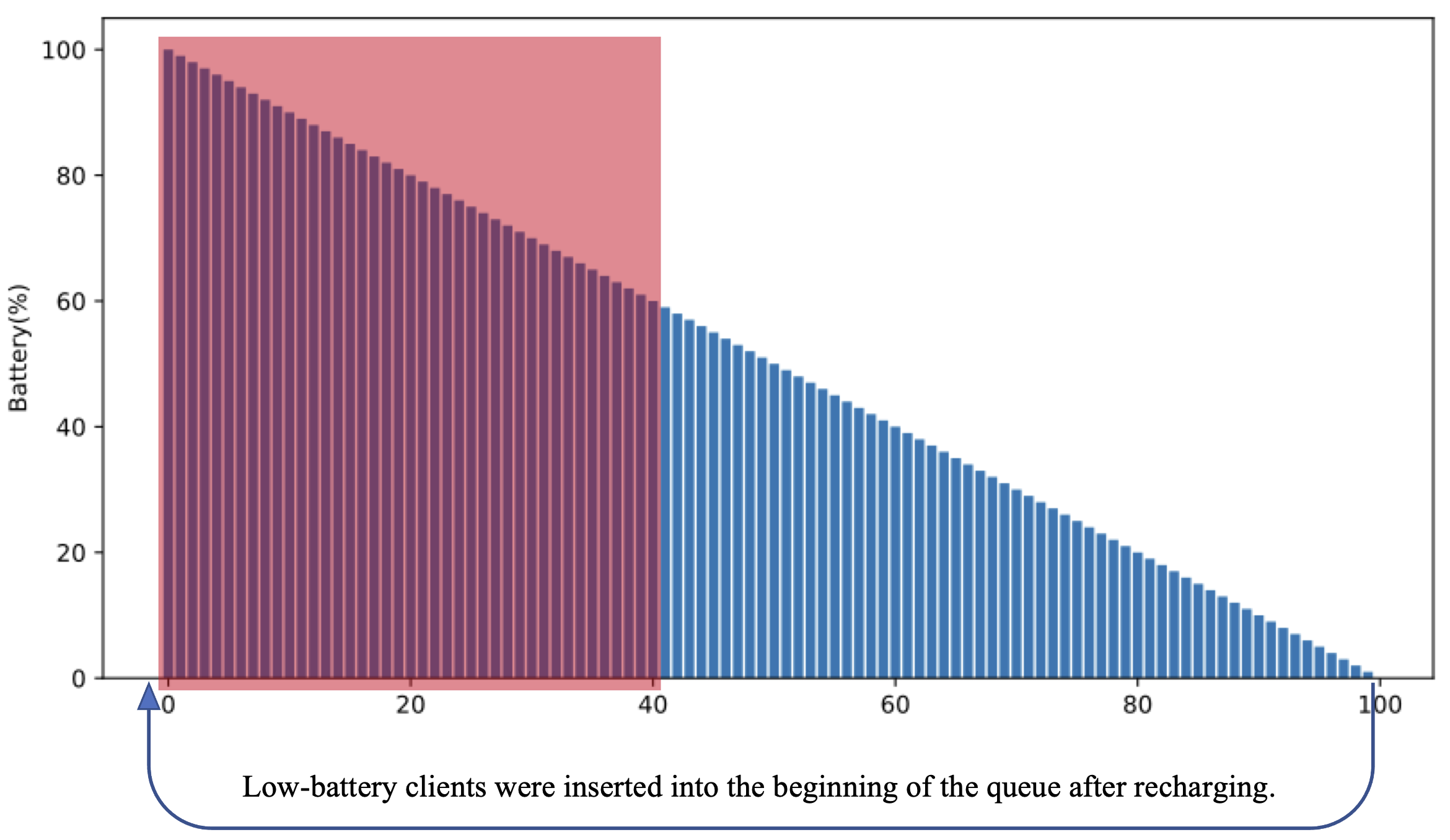}
    \caption{Only select devices with enough battery percentage.}
    \label{fig:battery_queue}
\end{figure}

In this paper, we illustrated how heterogeneity impacts the class imbalance issue, hence leading to a degraded performance in ASNs-based FL. Specifically, we visualize the relationship between heterogeneity and class imbalance of grouped data. We demonstrate how the battery life constraint exacerbates the class imbalance issue from two perspectives: (a) by limiting the number of devices available for FL training, and (b) by restricting the selection of devices to a smaller pool, as shown in Fig. \ref{fig:battery_queue}.
We conclude that ASNs-based FL is experiencing more severe class imbalance issues even under the same degree of heterogeneity. Finally, we study how different degrees of heterogeneity affect the FL training and the performance of current state-of-the-art algorithms on different degrees of heterogeneity.

\section{Related Work}
FL \citep{mcmahan2017communication} presents a promising avenue for training models that require substantial data volumes, all without the necessity of centralizing client data. Instead of transmitting raw data, FL employs a process where model parameters are communicated with edge devices during training. This method circumvents the significant communication overhead while upholding user privacy. While FL facilitates privacy-preserving distributed machine learning across myriad devices, it contends with persistent challenges, such as heterogeneity, within current methodologies. Heterogeneity manifests in various forms throughout FL training, adding complexity to the process. Additionally, the issue of class imbalance presents another formidable hurdle for FL, particularly when compounded with heterogeneity.

There are different types of heterogeneity issues, including statistical heterogeneity, system heterogeneity, communication heterogeneity, etc \citep{ye2023heterogeneous}. 
Statistical heterogeneity is mostly caused by the fact that the distributions vary among different clients, including label distribution and feature distribution, which causes the local models to converge towards different directions and the global model to converge slowly.
System heterogeneity mainly refers to the differences in hardware, network connectivity, and computational power among devices participating in federated learning.
Communication heterogeneity is variations in network speed and bandwidth across devices that can affect the efficiency of model transmission in federated learning, because of which, straggles may occur. There is already some research like \citep{li2020federated,reisizadeh2022straggler} focusing on tackling the straggler issue in FL to improve the overall performance.

\begin{figure*}[htbp]
    \centering
    \includegraphics[width=0.9\linewidth]{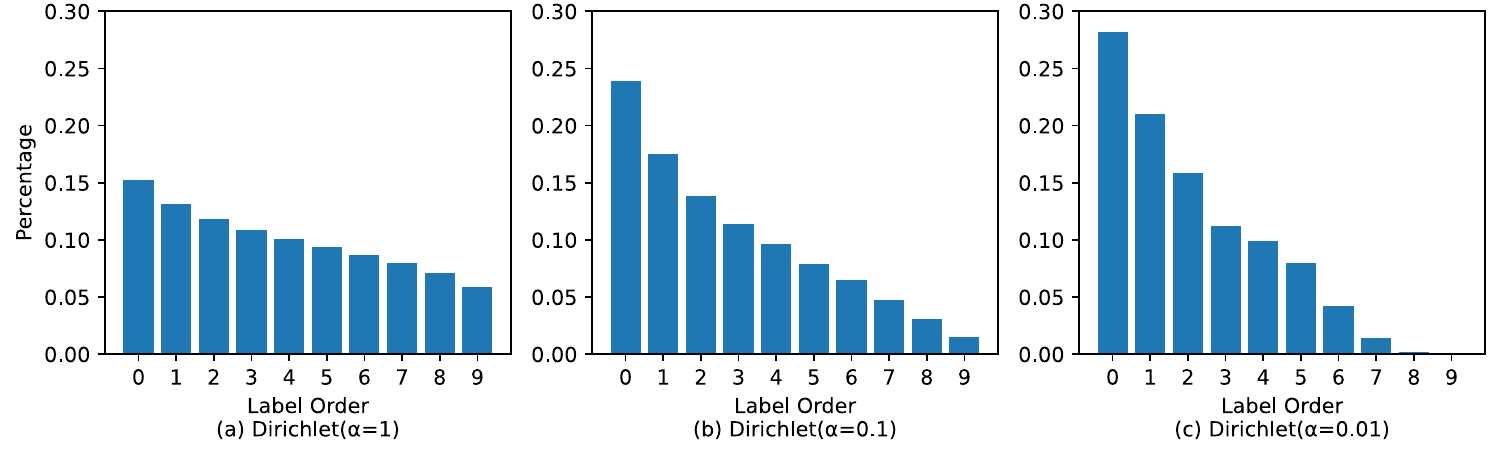}
    \caption{Grouped dataset (10 devices selected out of 100 devices) imbalance degree ($\alpha$) with various heterogeneity degrees under Dirichlet distribution. The smaller $\alpha$ is, the more heterogeneous the distribution among devices will be.}
    \label{fig:Imbalance_from_heterogeneity}
\end{figure*}

In this paper, we focus more on the statistical heterogeneity.
The impact of statistical heterogeneity was theoretically analyzed in \citep{li2019convergence}.
Plenty of research has been done to mitigate the impact of heterogeneity. 
FedProx \citep{li2020federated} introduced an additional proximal term to the local objection to refraining from overfitting local training. Despite that tuning the hyperparameter $\mu$ in FedProx could be a challenge, the introduced proximal term may also slow the convergence speed. FedProx is also capable of tackling the stragglers caused by communication heterogeneity.
Scaffold \citep{karimireddy2020scaffold} maintained control variates to rectify the local training to mitigate the heterogeneity effect. However, as shown in our experiments in Section V, Scaffold fails to outperform FedAvg \citep{mcmahan2017communication} when the heterogeneity degree is too high. 
FedMix \citep{yoon2021fedmix} relaxes the limitation of accessing others’ raw data and performs data augmentation with the assistance of other clients' data. By this strategy, FedMix could accommodate FL with different levels of privacy depending on applications and achieve better performance. 
MOON \citep{li2021model} used the similarity between model representations to correct for local training. On the contrary, relying on previous local models reduced its effectiveness when selecting a small portion of clients from a vast pool. 
WeiAvg \citep{dong2023weiavg} adopted weighted averaging to highlight updates from high-diversity clients under the diversity heterogeneity distribution. However, this will not work when the heterogeneity does not lie in diversity.
Despite these efforts, there is still significant room for improvement in addressing heterogeneity. While adding additional regularization terms \citep{li2020federated, li2021model} to local objective functions requires longer computation time. Algorithms like Scaffold \citep{karimireddy2020scaffold} could not even outperform FedAvg under highly heterogeneous distribution.

Class imbalance has long been an issue in the field of machine learning \citep{japkowicz2002class}. Resampling, re-weighting, and cost modification methods \citep{japkowicz2002class, cui2019class} have been proposed to mitigate its detriment. However, these techniques could not be applied to FL directly. Recently, \citep{zhang2023fed} pointed out that the class imbalance is the cause of performance degradation under non-IID settings.

\section{Heterogeneity and Imbalance}

Heterogeneity, specifically statistical heterogeneity, is caused by the distribution discrepancy among participating devices. This discrepancy leads to the grouped dataset's imbalance. It was noted in \citep{zhang2023fed} that the imbalance of the grouped dataset in FL leads to the degradation of model performance.

For a classification problem, suppose there are $B$ classes. For each global round, we select a set of devices to participate in the FL training. We can group the dataset across these selected devices together to obtain the grouped dataset in each global round. The distribution of the grouped dataset will be like $p = [p_1, p_2, \cdots, p_B]$. The imbalance degree of the grouped dataset could be represented as $\Delta = max(p) - min(p)$.

To examine the effects of varying degrees of heterogeneity on the imbalance of a grouped dataset, we simulate different distributions using the Dirichlet distribution. The $\alpha$ hyperparameter in the Dirichlet distribution controls the degree of heterogeneity. Lower values of $\alpha$ result in more heterogeneous distributions. We select 10 devices out of 100 clients then group the dataset together and repeat this process 100 times to get a robust average result. As shown in Fig. \ref{fig:Imbalance_from_heterogeneity}, the more heterogeneous the distribution is, the more imbalanced the grouped dataset will be. The imbalance degree $\Delta$ of the grouped dataset is 0.0897, 0.2117, and 0.2589 for $\alpha$ being 1, 0.1, and 0.01 respectively.

However, heterogeneity is not the only issue that could cause the imbalance among grouped datasets. In the next section, we will explain how device selection could also be a perpetrator.

\begin{figure}[htbp]
    \centering
    \includegraphics[width=0.9\linewidth]{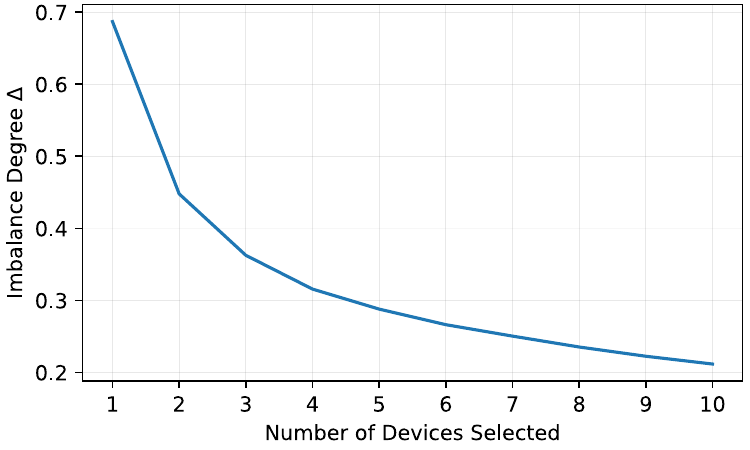}
    \caption{Grouped dataset imbalance degree with various number of clients select.}
    \label{fig:ImbalanceDegree}
\end{figure}

\begin{figure}[htbp]
    \centering
    \includegraphics[width=0.9\linewidth]{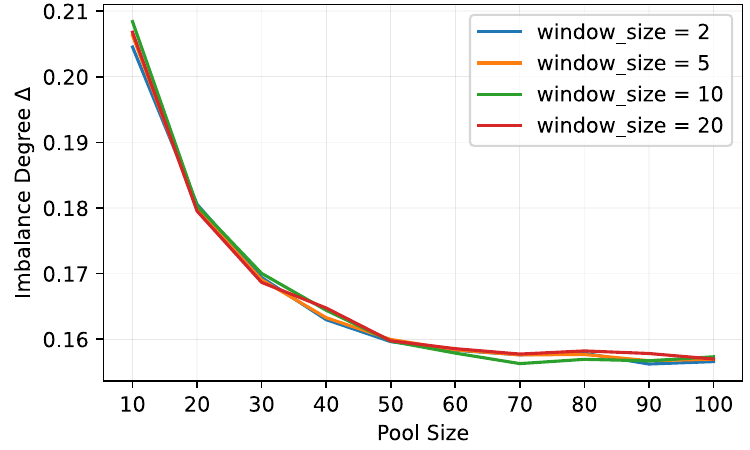}
    \caption{Imbalance degree under different pool sizes.}
    \label{fig:ImbalanceDegree_PoolSize}
\end{figure}

\begin{figure*}[htbp]
    \centering
    \includegraphics[width=0.9\linewidth]{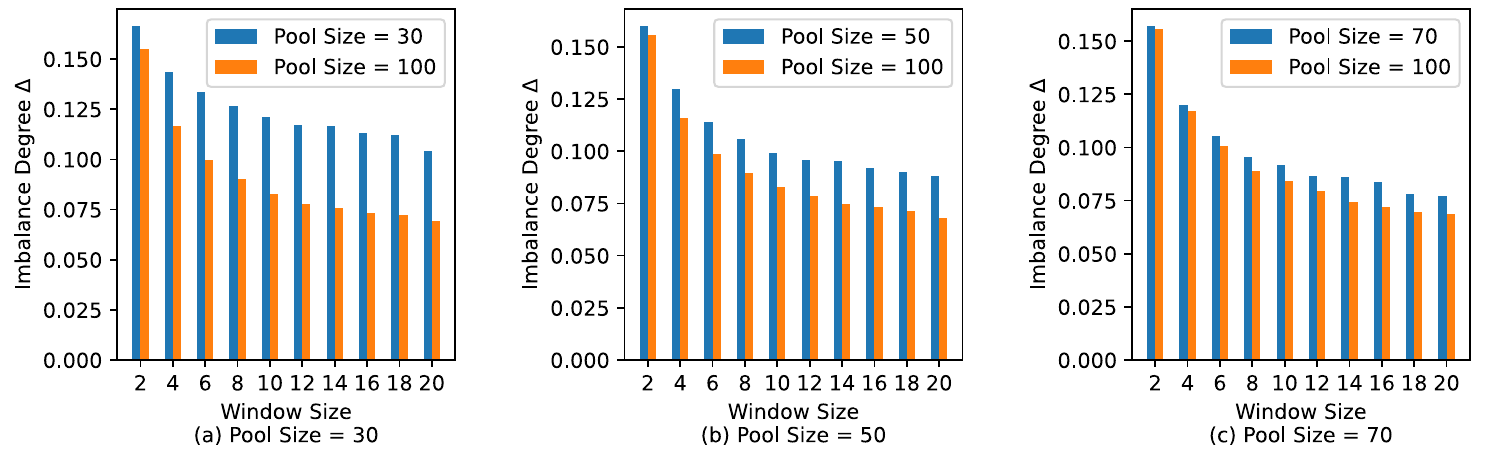}
    \caption{Imbalance degree under different window sizes with different pool sizes.}
    \label{fig:ImbalanceDegree_WindowSize}
\end{figure*}

\begin{figure*}[htbp]
    \centering
    \includegraphics[width=0.9\linewidth]{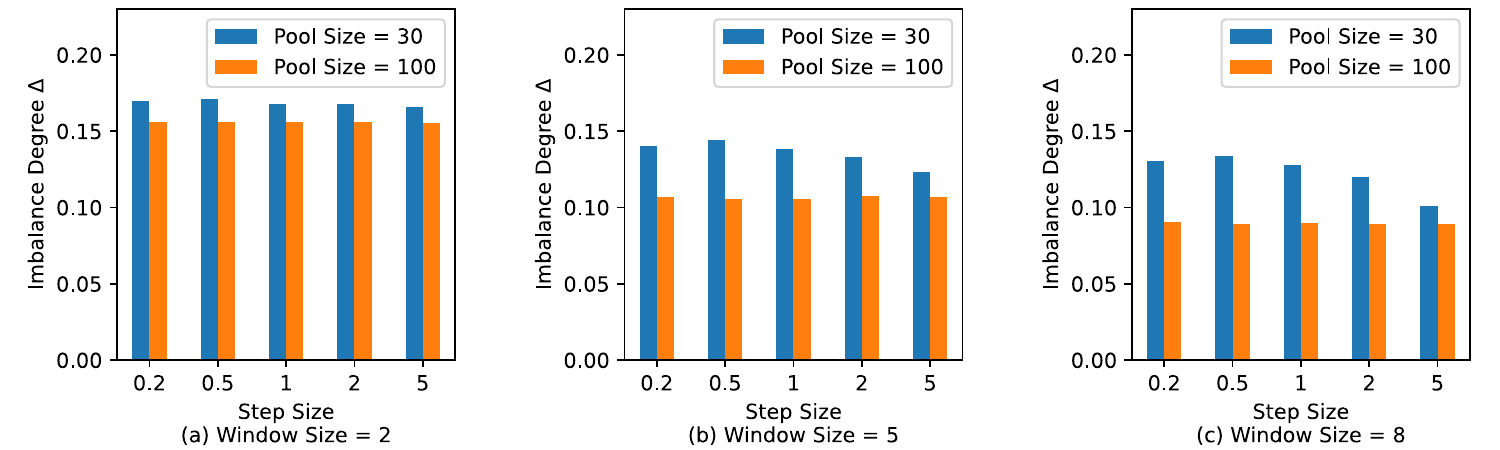}
    \caption{Imbalance degree under different step sizes with different window sizes.}
    \label{fig:ImbalanceDegree_StepSize}
\end{figure*}

\section{How Battery Life Constraint Aggravates the Imbalance Issue}
Compared with other IoT scenarios, FL training with aerial and space devices is often constrained by the battery life. In the near future, there is not likely to be a big step in terms of the energy density of lithium batteries \citep{fang2021challenges}. For example, the maximum flight time for a typical DJI drone is around 30 minutes. 
And the first priority of these devices should be finishing specific tasks or returning back.
This constraint will limit the choices when we select devices to participate in the FL training. The impact is twofold. On one hand, we may only be able to select fewer devices compared to other FL scenarios. On the other hand, we can only choose devices from a subset of all available devices, as those with low battery levels need to prioritize basic operations such as returning.

\subsection{Selecting fewer devices}
To investigate the impact of varying device sample sizes on imbalance patterns, we conducted experiments simulating different numbers of devices being selected. Each time we select a certain amount of devices, we group their local data together to see how imbalanced the grouped dataset is. We use the same distribution for different settings. We repeat 10,000 times and average the test results for a robust conclusion.
According to Fig. \ref{fig:ImbalanceDegree}, there is a clear trend showing that when fewer devices are selected in each round, the imbalance issue will aggravate.

\subsection{Selecting devices from a smaller pool}

Since the priority of aerial and space devices should be maintaining their normal operations, so once the battery percentage of some devices is below some threshold, we should not select them for FL training before they are recharged. 
In our training process, we can maintain a queue according to each device's battery percentage. We only select those top devices with the highest battery percentage. Those low-battery devices will also be recharged after a while and will enter the queue, as shown in Fig. \ref{fig:battery_queue}. Under this setting, our choice would be limited to a smaller amount of devices when selecting devices to participate in the local training. However, after the low-battery devices are recharged, the choice pool will also be updated.
To see more details of the imbalance issue, we merge the grouped dataset across several global rounds together to see the imbalance degree. Because we hope even if the ratio of a certain type of label is small in this global round, this situation will not last. 
As shown in Fig. \ref{fig:ImbalanceDegree_PoolSize}, when we can only select devices from a smaller device pool, the imbalance issue will aggravate, no matter what window size we choose.

\begin{figure*}[htbp]
    \centering
    \includegraphics[width=0.9\linewidth]{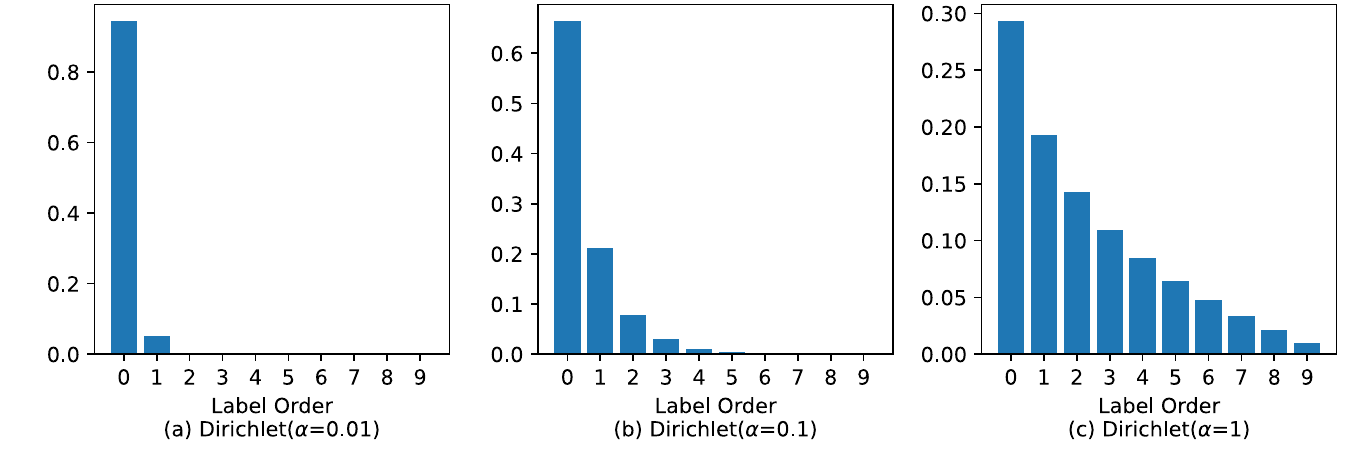}
    \caption{Average distribution on each client with different settings.}
    \label{fig:Distribution}
\end{figure*}

We further test the impact of the different choices of window sizes on imbalance degree $\Delta$.
We choose the pool size to be 30, 50, and 70. As shown in Fig. \ref{fig:ImbalanceDegree_WindowSize}, a smaller device pool consistently aggravates the imbalance issue under different observation window sizes.

Charging speed or updating speed of the pool will also impact the imbalance degree. In the above analysis, we adopted the same updating speed, which is in each global round, which will be one device entering the available pool. We test different updating speeds with each round updating 0.2, 0.5, 1, 2, and 5 devices respectively. As shown in Fig. \ref{fig:ImbalanceDegree_StepSize}, a smaller pool size consistently increases the imbalance degree of the grouped dataset across different observation window sizes.

With the analysis above, we can conclude that the battery constraint of aerial and space devices aggravates the imbalance issue. This exacerbates the existing heterogeneity problem even further. While previous research also studies the heterogeneity issue \citep{karimireddy2020scaffold,li2021model,cho2020client,yu2023latency}, the distributions they use are often not heterogeneous enough. In the next section, we study the impact of different heterogeneity degrees on FL training.


\begin{table*}[htbp]
\centering
\caption{Experiments of CIFAR10 dataset under Dirichlet distribution with different $\alpha$.}
\begin{tabular}{ccccccc}
\toprule
\multirow{2}{*}{Algorithms} & \multicolumn{2}{c}{Dirichlet($\alpha=1$)} & \multicolumn{2}{c}{Dirichlet($\alpha=0.1$)} & \multicolumn{2}{c}{Dirichlet($\alpha=0.01$)} \\ \cline{2-7} 
                            & TestAcc(std)            & Rounds          & TestAcc(std)             & Rounds           & TestAcc(std)              & Rounds           \\ \midrule
FedAvg                      & 0.6315(0.0051)          & 380             & 0.5936(0.0248)           & 418              & 0.3966(0.0578)            & 490              \\
FedProx                     & 0.6310(0.0051)          & 450             & 0.5943(0.0218)           & 418              & 0.4193(0.0536)            & 417              \\
MOON                        & 0.6288(0.0073)          & 383             & 0.5896(0.0232)           & -                & 0.3847(0.0710)            & -                \\
Scaffold                    & 0.6977(0.0048)          & 59              & 0.6462(0.0153)           & 153              & 0.2346(0.0828)            & -                \\ \bottomrule
\end{tabular}
\label{CIFAR10_Dirichlet}
\end{table*}

\section{How Degree of Heterogeneity Affects FL Training}

As observed in the paper \citep{zhang2023fed}, the essential reason resulting in FL performance degradation is the
class imbalance of the grouped dataset. Since different levels of heterogeneity degree $\alpha$ result in different levels of imbalance degree $\Delta$, hence resulting in different performance of the FL model, we can skip the intermediate steps and directly study the relationship between heterogeneity degree and FL model performance.

As shown in Fig. \ref{fig:Distribution}, we visualize how $\alpha$ affects the level of heterogeneity. We derive the visualization by following procedures: We sort each client's label distribution in descending order at first. Then we average the sorted label distribution among the clients to get a more stable result. So the more heterogeneous the distribution is, the more the barplot would be skewed.
When $\alpha$ equals 0.01, each client almost contains only one class of samples, as shown in Fig. \ref{fig:Distribution} (a). Note that the x-ticks do not denote the exact class label since we sort each client's distribution based on frequency in descending order before averaging them. With $\alpha$ increasing, the distribution becomes more uniform among all class labels.

\subsection{FedAvg Performance Under Different Heterogeneity Degrees}

\begin{figure}[htbp]
    \centering
    \includegraphics[width=0.9\linewidth]{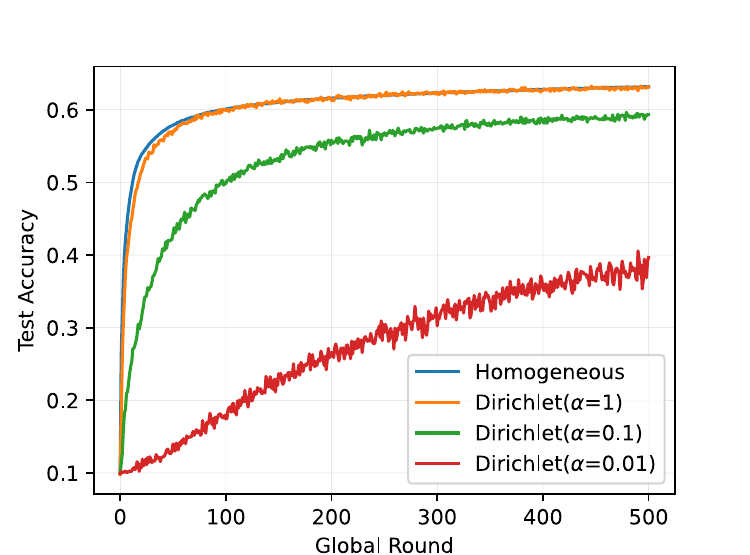}
    \caption{How the test accuracy lines look like under different levels of heterogeneity.}
    \label{fig:homogeneous}
\end{figure}

We demonstrate how different degrees of heterogeneity would affect FL training in the CIFAR10 dataset as shown in Fig. \ref{fig:homogeneous}. We set $\alpha$ to $10^{15}$ to simulate the homogeneous distribution, where each class exactly accounts for $10\%$ of the samples. All the experiments in Fig. \ref{fig:homogeneous} are based on the FedAvg \citep{mcmahan2017communication} algorithm. The only difference is the label distribution of each client. We conduct each experiment with different random seeds, then average them together to obtain a smoother and more solid test accuracy line. From Fig. \ref{fig:homogeneous}, we can derive that under the Dirichlet distribution with $\alpha=1$, the test accuracy line is very close to that of a homogeneous distribution. 
While other research often takes $\alpha = 0.1$ as the indication of heterogeneity, we show that $\alpha = 0.1$ is far from heterogeneous enough compared with $\alpha = 0.01$.
In the following part, we may denote the Dirichlet distribution with $\alpha=1$ as close to homogeneity, $\alpha=0.1$ as low heterogeneity, and $\alpha=0.01$ as high heterogeneity.

\subsection{State-of-the-art FL Algorithms Performance Under Different Heterogeneity Degrees}

We run several state-of-the-art algorithms under different degrees of heterogeneity. As shown in the experiment results in Table \ref{CIFAR10_Dirichlet}, FedProx \citep{li2020federated} could only marginally improve the performance, while MOON is even beaten by FedAvg \citep{mcmahan2017communication}. Scaffold \citep{karimireddy2020scaffold} performs pretty well under homogeneous and low-heterogeneity distribution but fails to outperform FedAvg \citep{mcmahan2017communication} under high heterogeneity.

\section{Conclusion}

In this paper, we identify a specific constraint of ASNs-based FL compared with other scenarios, which is the battery constraint. We then analyzed the impact of the battery constraint on FL training. We point out that the battery constraint will aggravate the heterogeneity and class imbalance issues from various perspectives, hence necessitating the FL optimization under high heterogeneity. Finally, we demonstrate that current state-of-the-art algorithms can not perform well under high heterogeneity. In future research, we will focus on FL optimization under highly heterogeneous distribution.

\bibliography{ref}

\begin{thebibliography}{17}
\providecommand{\natexlab}[1]{#1}
\providecommand{\url}[1]{#1}
\csname url@samestyle\endcsname
\providecommand{\newblock}{\relax}
\providecommand{\bibinfo}[2]{#2}
\providecommand{\BIBentrySTDinterwordspacing}{\spaceskip=0pt\relax}
\providecommand{\BIBentryALTinterwordstretchfactor}{4}
\providecommand{\BIBentryALTinterwordspacing}{\spaceskip=\fontdimen2\font plus
\BIBentryALTinterwordstretchfactor\fontdimen3\font minus \fontdimen4\font\relax}
\providecommand{\BIBforeignlanguage}[2]{{%
\expandafter\ifx\csname l@#1\endcsname\relax
\typeout{** WARNING: IEEEtranN.bst: No hyphenation pattern has been}%
\typeout{** loaded for the language `#1'. Using the pattern for}%
\typeout{** the default language instead.}%
\else
\language=\csname l@#1\endcsname
\fi
#2}}
\providecommand{\BIBdecl}{\relax}
\BIBdecl

\bibitem[Liu et~al.(2018)Liu, Shi, Fadlullah, and Kato]{liu2018space}
J.~Liu, Y.~Shi, Z.~M. Fadlullah, and N.~Kato, ``Space-air-ground integrated network: A survey,'' \emph{IEEE Communications Surveys \& Tutorials}, vol.~20, no.~4, pp. 2714--2741, 2018.

\bibitem[Zhang et~al.(2022)Zhang, Chen, Liu, Lan, Jiang, and Wan]{zhang2022aerial}
Y.~Zhang, C.~Chen, L.~Liu, D.~Lan, H.~Jiang, and S.~Wan, ``Aerial edge computing on orbit: A task offloading and allocation scheme,'' \emph{IEEE Transactions on Network Science and Engineering}, vol.~10, no.~1, pp. 275--285, 2022.

\bibitem[McMahan et~al.(2017)McMahan, Moore, Ramage, Hampson, and y~Arcas]{mcmahan2017communication}
B.~McMahan, E.~Moore, D.~Ramage, S.~Hampson, and B.~A. y~Arcas, ``Communication-efficient learning of deep networks from decentralized data,'' in \emph{Artificial intelligence and statistics}.\hskip 1em plus 0.5em minus 0.4em\relax PMLR, 2017, pp. 1273--1282.

\bibitem[Ye et~al.(2023)Ye, Fang, Du, Yuen, and Tao]{ye2023heterogeneous}
M.~Ye, X.~Fang, B.~Du, P.~C. Yuen, and D.~Tao, ``Heterogeneous federated learning: State-of-the-art and research challenges,'' \emph{ACM Computing Surveys}, vol.~56, no.~3, pp. 1--44, 2023.

\bibitem[Li et~al.(2020)Li, Sahu, Zaheer, Sanjabi, Talwalkar, and Smith]{li2020federated}
T.~Li, A.~K. Sahu, M.~Zaheer, M.~Sanjabi, A.~Talwalkar, and V.~Smith, ``Federated optimization in heterogeneous networks,'' \emph{Proceedings of Machine learning and systems}, vol.~2, pp. 429--450, 2020.

\bibitem[Reisizadeh et~al.(2022)Reisizadeh, Tziotis, Hassani, Mokhtari, and Pedarsani]{reisizadeh2022straggler}
A.~Reisizadeh, I.~Tziotis, H.~Hassani, A.~Mokhtari, and R.~Pedarsani, ``Straggler-resilient federated learning: Leveraging the interplay between statistical accuracy and system heterogeneity,'' \emph{IEEE Journal on Selected Areas in Information Theory}, vol.~3, no.~2, pp. 197--205, 2022.

\bibitem[Li et~al.(2019)Li, Huang, Yang, Wang, and Zhang]{li2019convergence}
X.~Li, K.~Huang, W.~Yang, S.~Wang, and Z.~Zhang, ``On the convergence of fedavg on non-iid data,'' \emph{arXiv preprint arXiv:1907.02189}, 2019.

\bibitem[Karimireddy et~al.(2020)Karimireddy, Kale, Mohri, Reddi, Stich, and Suresh]{karimireddy2020scaffold}
S.~P. Karimireddy, S.~Kale, M.~Mohri, S.~Reddi, S.~Stich, and A.~T. Suresh, ``Scaffold: Stochastic controlled averaging for federated learning,'' in \emph{International conference on machine learning}.\hskip 1em plus 0.5em minus 0.4em\relax PMLR, 2020, pp. 5132--5143.

\bibitem[Yoon et~al.(2021)Yoon, Shin, Hwang, and Yang]{yoon2021fedmix}
T.~Yoon, S.~Shin, S.~J. Hwang, and E.~Yang, ``Fedmix: Approximation of mixup under mean augmented federated learning,'' \emph{arXiv preprint arXiv:2107.00233}, 2021.

\bibitem[Li et~al.(2021)Li, He, and Song]{li2021model}
Q.~Li, B.~He, and D.~Song, ``Model-contrastive federated learning,'' in \emph{Proceedings of the IEEE/CVF conference on computer vision and pattern recognition}, 2021, pp. 10\,713--10\,722.

\bibitem[Dong et~al.(2023)Dong, Abbasi, Drew, Leung, Wang, and Zhou]{dong2023weiavg}
F.~Dong, A.~Abbasi, S.~Drew, H.~Leung, X.~Wang, and J.~Zhou, ``Weiavg: Federated learning model aggregation promoting data diversity,'' \emph{arXiv preprint arXiv:2305.16351}, 2023.

\bibitem[Japkowicz and Stephen(2002)]{japkowicz2002class}
N.~Japkowicz and S.~Stephen, ``The class imbalance problem: A systematic study,'' \emph{Intelligent data analysis}, vol.~6, no.~5, pp. 429--449, 2002.

\bibitem[Cui et~al.(2019)Cui, Jia, Lin, Song, and Belongie]{cui2019class}
Y.~Cui, M.~Jia, T.-Y. Lin, Y.~Song, and S.~Belongie, ``Class-balanced loss based on effective number of samples,'' in \emph{Proceedings of the IEEE/CVF conference on computer vision and pattern recognition}, 2019, pp. 9268--9277.

\bibitem[Zhang et~al.(2023)Zhang, Li, Tang, Sun, Chen, Zhang, Chen, Chen, and Li]{zhang2023fed}
J.~Zhang, A.~Li, M.~Tang, J.~Sun, X.~Chen, F.~Zhang, C.~Chen, Y.~Chen, and H.~Li, ``Fed-cbs: A heterogeneity-aware client sampling mechanism for federated learning via class-imbalance reduction,'' in \emph{International Conference on Machine Learning}.\hskip 1em plus 0.5em minus 0.4em\relax PMLR, 2023, pp. 41\,354--41\,381.

\bibitem[Fang(2021)]{fang2021challenges}
H.~Fang, ``Challenges with the ultimate energy density with li-ion batteries,'' in \emph{IOP Conference Series: Earth and Environmental Science}, vol. 781, no.~4.\hskip 1em plus 0.5em minus 0.4em\relax IOP Publishing, 2021, p. 042023.

\bibitem[Cho et~al.(2020)Cho, Wang, and Joshi]{cho2020client}
Y.~J. Cho, J.~Wang, and G.~Joshi, ``Client selection in federated learning: Convergence analysis and power-of-choice selection strategies,'' \emph{arXiv preprint arXiv:2010.01243}, 2020.

\bibitem[Yu et~al.(2023)Yu, Sun, Albelaihi, and Yi]{yu2023latency}
L.~Yu, X.~Sun, R.~Albelaihi, and C.~Yi, ``Latency-aware semi-synchronous client selection and model aggregation for wireless federated learning,'' \emph{Future Internet}, vol.~15, no.~11, p. 352, 2023.

\end{thebibliography}

\end{document}